\documentclass[runningheads]{llncs}
\usepackage{graphicx}
\usepackage{tabulary}
\usepackage{multirow}
\usepackage{pifont}
\usepackage{url}
\usepackage{xcolor}
\usepackage[super]{nth}
\usepackage{caption}
\usepackage{subcaption}
\usepackage{float}
\usepackage[utf8]{inputenc}
\usepackage[english]{babel}
\usepackage{amsmath}
\usepackage{hyperref}
\usepackage{multicol}
\usepackage{bbold}
\usepackage{amsfonts}
\usepackage{cite}
\usepackage[export]{adjustbox}
\usepackage{ragged2e} 
\usepackage{mathtools}
\usepackage{hyperref}
\newcommand{\defeq}{\vcentcolon=}

\definecolor{Gray}{gray}{0.9}
\definecolor{newcolor}{rgb}{.8,.349,.1}
\definecolor{Gray}{gray}{0.9}
\usepackage{soul} 

\newcommand{\rebuttal}[1]{\textcolor{black}{#1}}

\newcommand{\virg}[1]{``#1''}
\hypersetup{
    colorlinks=true,
    linkcolor=blue,
    filecolor=magenta,      
    urlcolor=blue,
}

\begin{document}
\title{Contrastive Learning with Continuous Proxy Meta-Data for 3D MRI Classification}
\titlerunning{Contrastive Learning with Continuous Proxy Meta-Data}
%
\author{Benoit Dufumier\inst{1,2}\and Pietro Gori\inst{2}\and Julie Victor\inst{1}\and Antoine Grigis\inst{1}\and
Michel Wessa\inst{3} \and
Paolo Brambilla\inst{4} \and
Pauline Favre\inst{1} \and
Mircea Polosan\inst{5} \and
Colm McDonald\inst{6} \and
Camille Marie Piguet \and
Edouard Duchesnay\inst{1} for the Alzheimer's Disease Neuroimaging Initiative }

\authorrunning{B. Dufumier et al.}
%
\institute{NeuroSpin, CEA Saclay, Université Paris-Saclay, France \\ \email{benoit.dufumier@cea.fr}\\
\and
LTCI, Télécom Paris, IPParis, France\\
\and 
Dep. of Neuropsychology,  Johannes-Gutenberg Univ. of Mainz, Germany
\and 
Dep. of Neurosciences, Fondazione IRCCS, University of Milan, Italy\\
\and
Université Grenoble Alpes, Inserm U1216, CHU Grenoble Alpe, France\\
\and
Centre for Neuroimaging \& Cognitive Genomics (NICOG), Galway, Ireland
}

\maketitle              
\begin{abstract}
Traditional supervised learning with deep neural networks requires a tremendous amount of labelled data to converge to a good solution. For 3D medical images, it is often impractical to build a large homogeneous annotated dataset for a specific pathology. Self-supervised methods offer a new way to learn a representation of the images in an unsupervised manner with a neural network. In particular, contrastive learning has shown great promises by (almost) matching the performance of fully-supervised CNN on vision tasks. Nonetheless, this method does not take advantage of available meta-data, such as participant's age, viewed as prior knowledge. Here, we propose to leverage continuous \textit{proxy} metadata, in the contrastive learning framework, by introducing a new loss called $y$-Aware InfoNCE loss. Specifically, we improve the positive sampling during pre-training by adding more positive examples with similar \textit{proxy} meta-data with the anchor, assuming they share similar discriminative semantic features.With our method, a 3D CNN model pre-trained on $10^4$ multi-site healthy brain MRI scans can extract relevant features for three classification tasks: schizophrenia, bipolar diagnosis and Alzheimer's detection. When fine-tuned, it also outperforms 3D CNN trained from scratch on these tasks, as well as state-of-the-art self-supervised methods. Our code is made publicly available \href{https://github.com/Duplums/yAwareContrastiveLearning}{here}.
\end{abstract}
\section{Introduction}

Recently, self-supervised representation learning methods have shown great promises, surpassing traditional transfer learning from ImageNet to 3D medical images \cite{zhou2020genesis}. These models can be trained without costly annotations and they offer a great initialization point for a wide set of downstream tasks, avoiding the domain gap between natural and medical images. They mainly rely on a pretext task that is informative about the prior we have on the data. This proxy task essentially consists in corrupting the data with non-linear transformations that preserve the semantic information about the images and learn the reverse mapping with a Convolutional Neural Network (CNN). Numerous tasks have been proposed both in the computer vision field (inpainting \cite{pathak2016inpainting}, localization of a patch \cite{doersch2015patchloc}, prediction of the angle of rotation \cite{gidaris2018rot}, jigsaw \cite{noroozi2016jigsaw}, etc.) and also specifically designed for 3D medical images (context restoration \cite{chen2019self_supervision}, solving the rubik's cube \cite{zhuang2019rubikscube}, sub-volumes deformation \cite{zhou2020genesis}). They have been successfully applied to 3D MR images for both segmentation and classification \cite{zhuang2019rubikscube, tao2020rubikscube, taleb20203d, zhou2020genesis}, outperforming the classical 2D approach with ImageNet pre-training.
Concurrently, there has been a tremendous interest in contrastive learning \cite{hadsell2006dimensionality} over the last year. Notably, this unsupervised approach almost matches the performance over fully-supervised vision tasks and it outperforms supervised pre-training \cite{chen2020simCLR, caron2020unsupervised, he2020moco}. 
A single encoder is trained to map semantically similar ``positive'' samples close together in the latent space while pushing away dissimilar ``negative'' examples. In practice, all samples in a batch are transformed twice through random transformations $t\sim \mathcal{T}$ from a set of parametric transformations $\mathcal{T}$. For a given reference point (anchor) $x$, the positive samples are the ones derived from $x$ while the other samples are considered as negatives. Most of the recent works focus in finding the best transformations $\mathcal{T}$ that degrade the initial image $x$ while preserving the semantic information \cite{chen2020simCLR, tian2020} and very recent studies intend to improve the negative sampling \cite{robinson2020negative, chuang2020debiased}. However, two different samples are not necessarily semantically different, as emphasized in \cite{chuang2020debiased,wei2020co2}, and they may even belong to the same semantic class. Additionally, two samples are not always equally semantically different from a given anchor and so they should not be equally distant in the latent space from this anchor.
In this work, we assume to have access to continuous \textit{proxy} meta-data containing relevant information about the images at hand (\textit{e.g} the participant's age). 
We want to leverage these $\textit{meta-data}$ during the contrastive learning process in order to build a more universal representation of our data. To do so, we propose a new $y$-Aware InfoNCE loss inspired from the Noise Contrastive Estimation loss \cite{gutmann2010infoNCE} that aims at improving the positive sampling according to the similarity between two \textit{proxy} meta-data. Differently from \cite{khosla2020}, i) we perform contrastive learning with continuous meta-data (not only categorical) and ii) our first purpose is to train a generic encoder that can be easily transferred to various 3D MRI target datasets for classification or regression problems in the very small data regime ($N \le 10^3$). It is also one of the first studies to apply contrastive learning to 3D anatomical brain images \cite{chaitanya2020contrastive}.
Our main contributions are:\\
- we propose a novel formulation for contrastive learning that leverages \textit{continuous} meta-data and derive a new loss, namely the $y$-Aware InfoNCE loss\\
- we empirically show that our unsupervised model pre-trained on a large-scale multi-site 3D brain MRI dataset comprising $N=10^4$ healthy scans reaches or outperforms the performance of CNN model fully-supervised on 3 classification tasks under the linear protocol evaluation\\
- we also demonstrate that our approach gives better results when fine-tuning on 3 target tasks than training from scratch\\
- we finally performed an ablation study showing that leveraging the meta-data improves the performance for all the downstream tasks and different set of transformations $\mathcal{T}$ compared to SimCLR \cite{chen2020simCLR}

\section{Method}

\begin{figure}
    \centering
    \includegraphics[width=10.5cm]{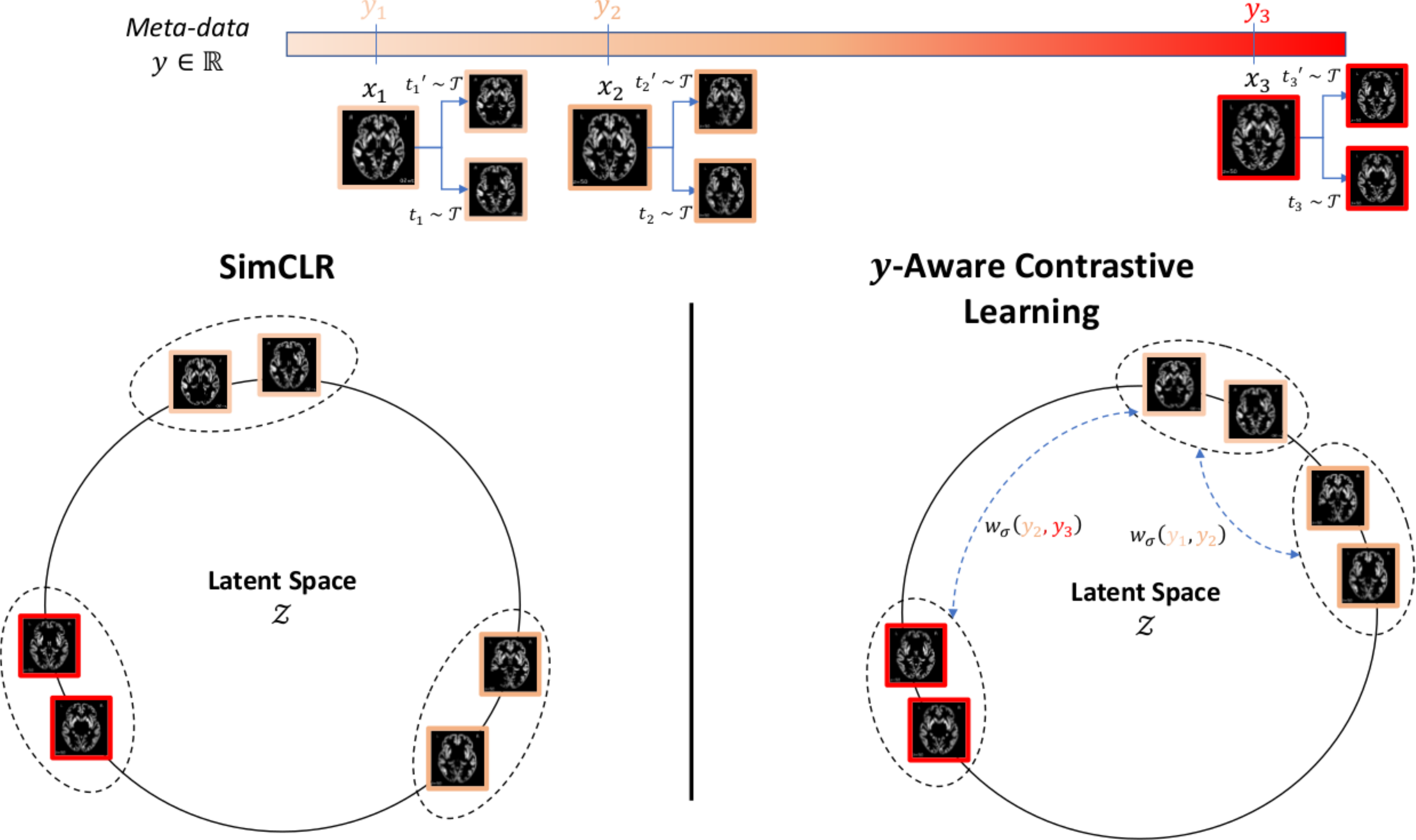}
    \caption{Differently from SimCLR \cite{chen2020simCLR}, our new loss can handle meta-data $y\in \mathbb{R}$ by redefining the notion of similarity between two images in the latent space $\mathcal{Z}$. For an image $x_i$, transformed twice through two augmentations $t_1, t_1'\sim \mathcal{T}$, the resulting views $(t_1(x_i), t_2(x_i))$ are expected to be close in the latent space  through the learnt mapping $f_\theta$, as in SimCLR. However, we also expect a different input $x_{k\neq i}$ to be close to $x_i$ in $\mathcal{Z}$ if the two \textit{proxy} meta-data $y_i$ and $y_k$ are similar. We define a similarity function $w_\sigma(y_i, y_k)$ that quantifies this notion of similarity.}
\end{figure}


\textbf{Problem Formalization.} In contrastive learning \cite{chen2020simCLR, khosla2020, tian2020}, one wants to learn a parametric function $f_\theta: \mathcal{X} \mapsto \mathbb{S}^{d-1}=\mathcal{Z}$ between the input image space $\mathcal{X}$ and the unit hypersphere, without meta-data. The goal of $f_\theta$ is to map samples to a representation space where semantically similar samples are \virg{closer} than semantically different samples. To do that, each training sample $x_i\in \mathcal{X}$ is transformed twice through $t_1^i, t_2^i\sim \mathcal{T}$ to produce two augmented views of the same image $(v_1^i, v_2^i)\defeq (t_1^i(x_i), t_2^i(x_i))$, where $\mathcal{T}$ is a set of predefined transformations. Then, for each sample $i$, the model $f_\theta$ is trained to discriminate between the \virg{positive} pair $(v_1^i, v_2^i)$, assumed to be drawn from the empirical joint distribution $p(v_1, v_2)$, and all the other \virg{negative} pairs $(v_1^i, v_2^j)_{j\neq i}$, assumed to be drawn from the marginal distributions $p(v_1)p(v_2)$. With these assumptions, $f_\theta$ is an estimator of the mutual information $I$ between $v_1$ and $v_2$ and it usually estimated by maximizing a lower bound of $I(v_1, v_2)$ called the InfoNCE loss \cite{oord2018representation}:


\begin{equation}
    \label{infoNCE_loss}
    \mathcal{L}_{NCE} = -\log \frac{e^{f_\theta(v_1^i, v_2^i)}}{\frac{1}{n}\sum_{j=1}^n e^{f_\theta(v_1^i, v_2^j)}}
\end{equation}
where $n$ is the batch size, $f_\theta(v_1, v_2) \defeq \frac{1}{\tau}f_\theta(v_1)^Tf_\theta(v_2)$ and $\tau > 0$ is a hyperparameter. $f_\theta$ is usually defined as the composition of an encoder network $e_{\theta_1}(x)$ and a projection head $z_{\theta_2}$(e.g. multi-layer perceptron) which is discarded after training (here $\theta=\{\theta_1, \theta_2\}$). Outputs lie on the unit hypersphere so that inner products can be used to measure cosine similarities in the representation space.  
In Eq. \ref{infoNCE_loss}, every sample $v_2^j|_{j\neq i}$ is considered \textit{equally} different from the anchor $v_1^i$. However, this is hardly true with medical images since we know, for instance, that two young healthy subjects should be considered more similar than a young and an old healthy subject. If we suppose to have access to continuous proxy metadata $y_i \in \mathbb{R}$ (\textit{e.g} participant's age or clinical score), then two views $v_1^i$, $v_2^j$ with similar metadata $y_i, y_j$ should be also close in the representation space $\mathbb{S}^{d-1}$. Inspired by vicinal risk minimization (VRM) \cite{chapelle2001vicinal}, we propose to re-define $p(v_1, v_2)$ by integrating the proxy metadata $y$, modeled as a random variable, such that a small change in $y$ results in a negligible change in $p(v_1, v_2 | y)$. Similarly to \cite{ding2020ccgan}, we define the empirical joint distribution as:

\begin{equation}
    p_{emp}^{vic}(v_1, v_2 | y) = \frac{1}{n}\sum_{i=1}^n\sum_{j=1}^n \frac{w_\sigma(y_i, y_j)}{\sum_{k=1}^n w_\sigma(y_i, y_k)} \delta(v_1 - v_1^i) \delta(v_2 - v_2^j)
    \label{approx_pos_distribution}
\end{equation}

where $\sigma > 0$ is the hyperparameter of the Radius Basis Function (RBF) kernel $w_\sigma$. 
Based on eq. \ref{approx_pos_distribution}, we can introduce our new $y$-Aware InfoNCE loss:
\begin{equation}
\label{y_aware_infoNCE_loss}
    \mathcal{L}^{y}_{NCE} = -\sum_{k=1}^{n}\frac{w_\sigma(y_k, y_i)}{\sum_{j=1}^n w_\sigma(y_j, y_i)}\log\frac{e^{f_\theta(v_1^i, v_2^k)}}{\frac{1}{n}\sum_{j=1}^n e^{f_\theta(v_1^i, v_2^j)}}
\end{equation}

In the limit case when $\sigma \rightarrow 0$, then we retrieve exactly the original InfoNCE loss, assuming that $y_i=y_j \Leftrightarrow x_i = x_j, \forall i, j \in [1..n]$. When $\sigma \rightarrow +\infty$, we assume that all samples $(x_i)_{i=1}^n$ belong to the same latent class.

\textbf{Discrete case.} If the proxy meta-data $(y_i)_{i\in [1..N]}$ are discrete, then we can simplify the above expression by imposing $w_\sigma(y_i, y_k) = \delta(y_i-y_k)$ retrieving the Supervised Contrastive Loss \cite{khosla2020}. We may see $\mathcal{L}^y_{NCE}$ as an extension of \cite{khosla2020} in the continuous case. However, our purpose here is to build a robust  encoder that can leverage meta-data to learn a more generalizable representation of the data.

\textbf{Generalization.} The proposed loss could be easily adapted to multiple metadata, both continuous and categorical, by defining one kernel per metadata. Other choices of kernel, instead of the RBF, could also be considered.



\textbf{Choice of the transformations $\mathcal{T}$.}
In our formulation, we did not specify particular transformations $\mathcal{T}$ to generate $(v_1^i, v_2^i)$. While there have been recent works \cite{chen2019self_supervision, tian2020} proposing transformations on natural images (color distorsion, cropping, cutout \cite{devries2017cutout}, etc.), there is currently no consensus for medical images in the context of contrastive learning. Here, we design three sets of transformations that preserve the semantic information in MR images: cutout, random cropping and a combination of the two with also gaussian noise, gaussian blur and flip. Importantly, while color distortion is crucial on natural images \cite{chen2020simCLR} to avoid the model using a shortcut during training based on the color histogram, it is not necessarily the case for MR images (see Supp. \ref{supp_color_histo}).

\section{Experiments}

\begin{figure}[h!]
\centering
\begin{subfigure}[b]{\textwidth}
   \includegraphics[width=\linewidth]{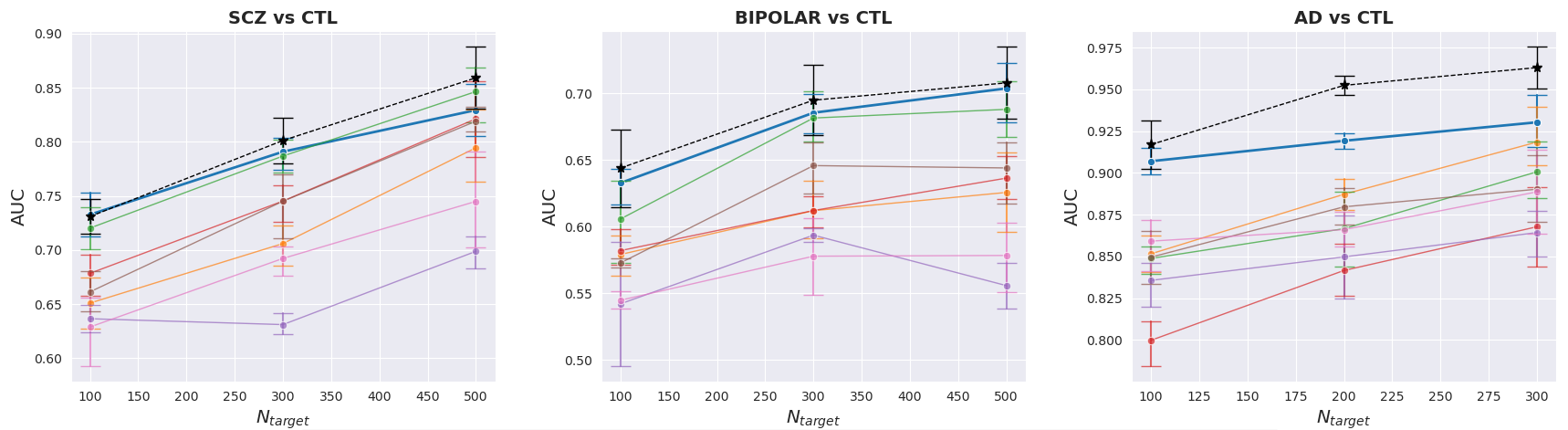}
   \caption{\rebuttal{5-fold CV Stratified on Site}}
\end{subfigure}

\begin{subfigure}[b]{\textwidth}
   \includegraphics[width=\linewidth]{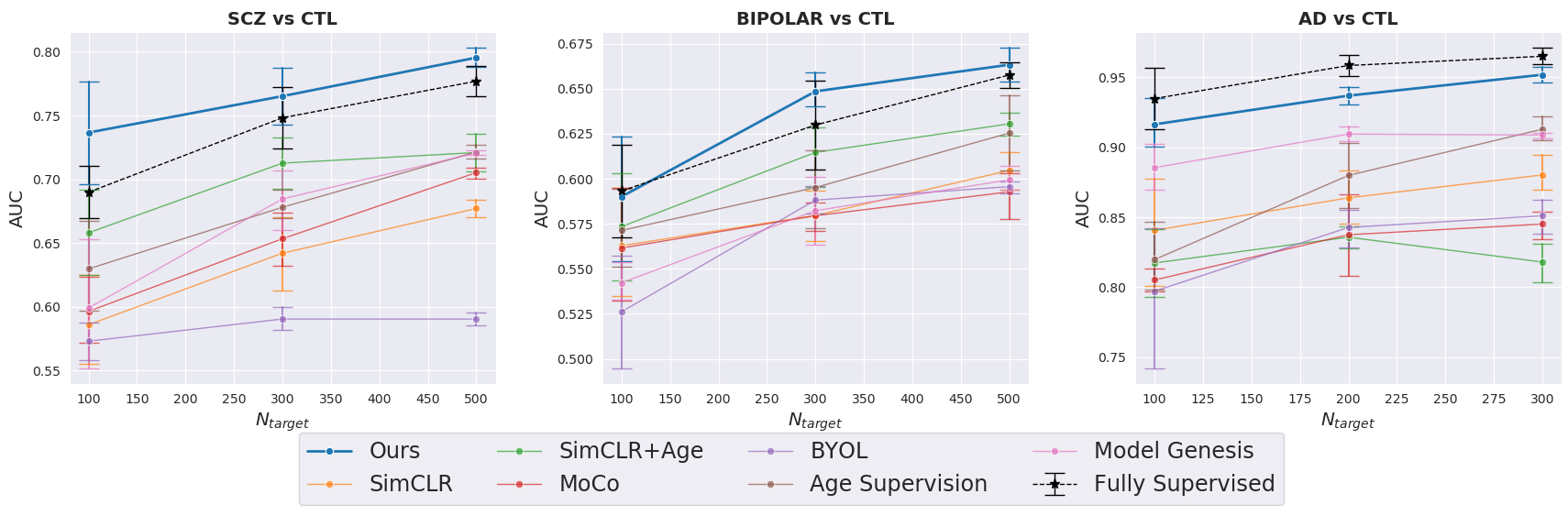}
   \caption{\rebuttal{5-fold CV Leave-Site-Out}}
\end{subfigure}
\caption{Comparison of different representations in terms of classification accuracy (downstream task) on three different data-sets (one per column). Classification is performed using a linear layer on top of the pre-trained frozen encoders. (a) Data for training/validation and test come from the the same acquisition sites (b) Data for training/validation and test come from different sites.} 

\label{EvaRepr}
\end{figure}

\underline{\textbf{Datasets}}\\ - \textbf{Big Healthy Brains (BHB) dataset} We aggregated 13 publicly available datasets\footnote{Demographic information as well as the public repositories can be found in Supp. \ref{supp_bhb}} of 3D T1 MRI scans of healthy controls (HC) acquired on more than 70 different scanners and comprising $N=10^4$ samples. We use this dataset only to pre-train our model with the \textbf{participant's age as the \textit{proxy} meta-data}. The learnt representation is then tested on the following \rebuttal{four} data-sets using as final task a binary classification between HC and patients.\\
- \textbf{SCHIZCONNECT-VIP\footnote{http://schizconnect.org}} It comprises $N=605$ multi-site MRI scans including 275 patients with strict schizophrenia (SCZ) and 330 HC.\\ 
- \textbf{BIOBD  \cite{hozer2020biobd, sarrazin2018}} This dataset includes $N=662$ MRI scans acquired on 8 different sites with 356 HC and 306 patients with bipolar disorder (BD).\\
- \textbf{BSNIP} \cite{tamminga2014bipolar} It includes $N=511$ MRI scans with $N=200$ HC, $N=194$ SCZ and $N=117$ BD. This independent dataset is used only at test time in Fig. \ref{EvaRepr}b). \\
- \textbf{Alzheimer's Disease Neuroimaging Initiative (ADNI-GO)\footnote{http://adni.loni.usc.edu/about/adni-go}} 
We use $N=387$ co-registered T1-weighted MRI images divided in $N=199$ healthy controls and $N=188$ Alzheimer's patients (AD). We only included one scan per patient at the first session (baseline).\\
All data-sets have been pre-processed in the same way with a non-linear registration to the MNI template and a gray matter extraction step. The final spatial resolution is $1.5mm$ isotropic and the images are of size $121\times 145\times 121$. 

\justify
\underline{\textbf{Implementation details}}
We implement our new loss based on the original InfoNCE loss \cite{chen2020simCLR} with Pytorch \cite{paszke2019pytorch} and we use  
the Adam optimizer during training. As opposed to SimCLR \cite{chen2020simCLR} and in line with \cite{chaitanya2020contrastive}, we only use a batch size of $b=64$ as it did not significantly change our results (see Supp. \ref{supp_hyperparams}). We also follow \cite{chen2020simCLR} by fixing $\tau=0.1$ in Eq.\ref{infoNCE_loss} and Eq.\ref{y_aware_infoNCE_loss} and we set the learning rate to $\alpha=10^{-4}$, decreasing it by $0.9$ every $10$ epochs. The model $e_{\theta_1}$ is based on a 3D adaptation of DenseNet121 \footnote{Detailed implementation in our \href{https://github.com/Duplums/yAwareContrastiveLearning}{repository}} \cite{DenseNet_Huang} and $z_{\theta_2}$ is a vanilla  multilayer perceptron as in \cite{chen2020simCLR}.





\justify
\underline{\textbf{Evaluation of the representation}}
In Fig.\ref{EvaRepr}, we compare the representation learnt using our model $f_\theta$ with the ones estimated using \rebuttal{i)} the InfoNCE loss (SimCLR) \cite{chen2020simCLR}, \rebuttal{ii)} Model Genesis \cite{zhou2020genesis}, a SOTA model for self-supervised learning with medical images, \rebuttal{iii)} a standard pre-training on age using a supervised approach (i.e. $l_1$ loss for age prediction), \rebuttal{iv) BYOL \cite{grill2020byol} and MoCo \cite{he2020moco} (memory bank $K=1024$) , 2 recently proposed SOTA models for representation learning, v) a multi-task approach SimCLR with age regression in the latent space (SimCLR+Age)} and a fully fine-tuned supervised DenseNet trained to predict the final task. This can be considered as an upper bound, if the training data-set is sufficiently big.
For the pre-training of our algorithm $f_\theta$, we only use the BHB dataset with the participant's age as \textit{proxy} meta-data. For both \rebuttal{contrastive learning methods} and BYOL, we fix $\sigma=5$ in Eq.\ref{y_aware_infoNCE_loss} and Eq.\ref{infoNCE_loss} and only use random cutout for the transformations $\mathcal{T}$ with a black patch covering $p=25\%$ of the input image. \rebuttal{We use UNet for pre-training with Model Genesis and DenseNet121 for all other models.}

In order to evaluate the quality of the learnt representations, we only added a linear layer on top of the frozen encoders pre-trained on BHB. We tune this linear layer on 3 different binary classification tasks (see Datasets section) with 5-fold cross-validation (CV). We tested two different situations: data for training/validation and test come either from the same sites (first row) or from different sites (second row). We also vary the size (i.e. number of subjects, $N_{\text{target}}$) of the training/validation set. For (a), we perform a stratified nested CV (two 5-fold CV, the inner one for choosing the best hyper-parameters and the outer one for estimating the test error). 
For (b), we use a 5-fold CV for estimating the best hyper-parameters and keep an independent constant test set for all $N_{target}$ (see Supp. \ref{supp_CV}).

From Fig. \ref{EvaRepr}, we notice that our method consistently outperforms the other pre-trainings even in the very small data regime ($N=100$) and it matches the performance of the fully-supervised setting on 2 data-sets. Differently from age supervision, $f_\theta$ is less specialized on a particular proxy task and it can be directly transferred on the final task at hand without fine-tuning the whole network. \rebuttal{Furthermore, compared to the multi-task approach SimCLR+Age, the features extracted by our method are less sensitive to the site where the MR images are coming from. This shows that our technique is the only one that efficiently uses the highly multi-centric dataset BHB by making the features learnt during pre-training less correlated to the acquisition sites.}

\begin{figure}[h]
    \centering
    \includegraphics[width=1\linewidth]{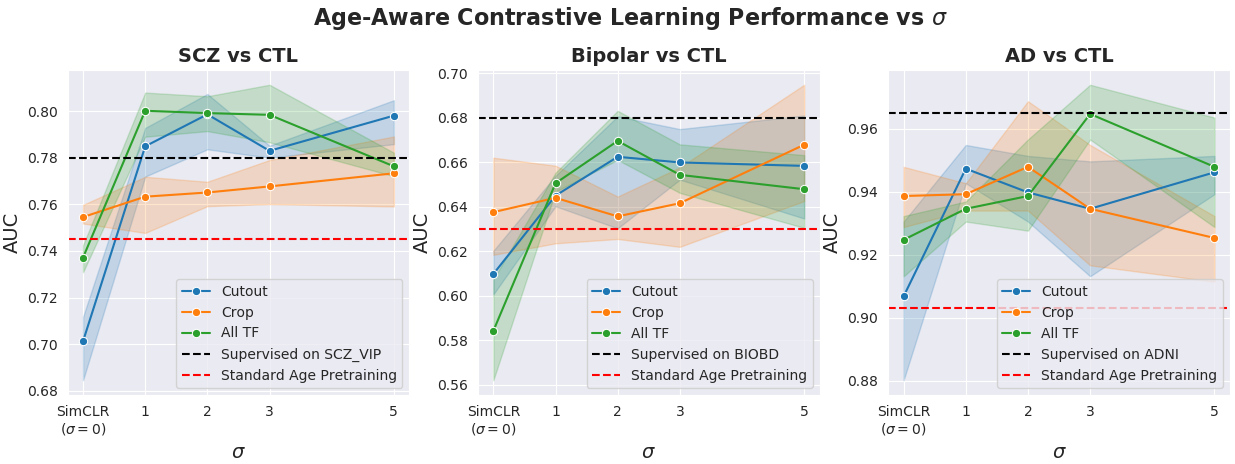}
    \caption{Linear classification performance on three binary classification tasks with $N_{pretrained}=10^4$. \textit{All TF} includes crop, cutout, gaussian noise, gaussian blur and flip. The encoder is frozen and we only tune a linear layer on top of it. $\sigma=0$ corresponds to SimCLR \cite{chen2020simCLR} with InfoNCE loss. As we increase $\sigma$, we add more positive examples for a given anchor $x_i$ with close proxy meta-data.}
    \label{ablation_study}
\end{figure}

\justify
\underline{\textbf{Importance of $\sigma$ and $\mathcal{T}$ in the positive sampling}}
In Fig. \ref{ablation_study}, we study the impact of $\sigma$ in Eq. \ref{y_aware_infoNCE_loss} on the final representation learnt for a given set of transformations $\mathcal{T}$. As highlighted in \cite{chen2020simCLR}, hard transformations seem to be important for contrastive learning (at least on natural images), therefore we have evaluated three different sets of trasformations $\mathcal{T}_1$ = $\{$ Random Crop $\}$, $\mathcal{T}_2$ = $\{$ Random Cutout $\}$ and $\mathcal{T}_3$ = $\{$ Cutout, Crop, Gaussian Noise, Gaussian Blur, Flip $\}$. Importantly, we did not include color distorsion in $\mathcal{T}_3$ since i) it is not adapted to MRI images where a voxel's intensity encodes a gray matter density and ii) we did not observe significant difference between the color histograms of different scans as opposed to \cite{chen2020simCLR} (see Supp. \ref{supp_color_histo}). As before, we evaluated our representation under the linear evaluation protocol. We can observe that $\mathcal{T}_1$ and $\mathcal{T}_3$ give similar performances with $\sigma > 0$, always outperforming both SimCLR ($\sigma=0$) and age supervision on BHB. It also even outperforms the fully-supervised baseline on SCZ vs HC. We also find that a strong cropping or cutout strategy is detrimental for the final performances (see Supp. \ref{supp_hyperparams}). Since $\mathcal{T}_1$ is computationally less expensive than $\mathcal{T}_3$, we chose to use $\mathcal{T}=\mathcal{T}_1$ and $\sigma=5$ in our experiments.

\justify
\underline{\textbf{Fine-tuning Results}}
Finally, we fine-tuned the whole encoder $f_\theta$ with different initializations on the 3 downstream tasks (see Table \ref{fine-tuning_table}). To be comparable with Model Genesis \cite{zhou2020genesis}, we also used the same UNet backbone for $f_\theta$ and we still fixed $\mathcal{T}_1 = \{\text{Random Cutout}\}$ and $\sigma=5$. First, our approach outperforms the CNNs trained from scratch on all tasks as well as Model Genesis, even with the same backbone. Second, when using DenseNet, our pre-training remains better than using age supervision as pre-training for SCZ vs HC (even with the same transformations) and it is competitive on  BD vs HC and AD vs HC.

\begin{table}[h!]
\resizebox{\linewidth}{!}{
    \centering
    \begin{tabular}{c|c|c c|c c|c c}
    \hline
    \multirow{2}{*}{\textbf{Backbone}} & \multirow{2}{*}{\textbf{Pre-training}} & \multicolumn{2}{c|}{\textbf{SCZ vs HC}} & \multicolumn{2}{c|}{\textbf{BD vs HC}} & \multicolumn{2}{c}{\textbf{AD vs HC}}\\
                                      & & \rebuttal{$N_{train}=100$} & \rebuttal{$N_{train}=500$} & \rebuttal{$N_{train}=100$} & \rebuttal{$N_{train}=500$}  & \rebuttal{$N_{train}=100$} & \rebuttal{$N_{train}=300$} \\
    \hline
    \multirow{4}{*}{UNet} & None & $72.62_{\pm0.9}$ & $76.45_{\pm2.2}$ & $63.03_{\pm2.7}$ & $69.20_{\pm3.7}$ & $88.12_{\pm3.2}$ & $94.16_{\pm3.9}$ \\
                          & Model Genesis \cite{zhou2020genesis} &$73.00_{\pm3.4}$ & $81.8_{\pm4.7}$ & $60.96_{\pm1.8}$ & $67.04_{\pm4.4}$ & $89.44_{\pm 2.6}$ & $95.16_{\pm3.3}$ \\
                          & SimCLR \cite{chen2019self_supervision}& $73.63_{\pm 2.4}$ & $80.12_{\pm 4.9}$ & $59.89_{\pm 2.6}$ & $66.51_{\pm4.3}$ & $90.60_{\pm2.5}$ & $94.21_{\pm2.7}$ \\
                          & \rebuttal{Age Prediction w/ D.A} & $\underline{75.32}_{\pm2.2}$ & $\underline{85.27}_{\pm2.3}$ & $\mathbf{64.6_{\pm1.6}}$ & $\mathbf{70.78_{\pm2.1}}$ &$\underline{91.71}_{\pm1.1}$ & $\underline{95.26}_{\pm1.5}$\\
                          & Age-Aware Contrastive Learning (ours) & $\textbf{75.95}_{\pm 2.7}$ & $\mathbf{85.73}_{\pm 4.7}$ & $\underline{63.79}_{\pm3.0}$ & $\underline{70.35}_{\pm2.7}$ & $\mathbf{92.19_{\pm1.8}}$ & $\mathbf{96.58_{\pm1.6}}$\\
    \hline
    \multirow{6}{*}{DenseNet} & None & $73.09_{\pm1.6}$ & $85.92_{\pm2.8}$ & $64.39_{\pm2.9}$ & $70.77_{\pm2.7}$& $92.23_{\pm1.6}$& $93.68_{\pm1.7}$ \\
                               & None w/ D.A & $\underline{74.71}_{\pm1.3}$ & $86.94_{\pm2.8}$ & $64.79_{\pm1.3}$ & $72.25_{\pm1.5}$& $92.10_{\pm 1.8}$ & $94.16_{\pm2.5}$ \\
                               & SimCLR \cite{chen2020simCLR} & $70.80_{\pm1.9}$ & $86.35_{\pm2.2}$ & $60.57_{\pm1.9}$ & $67.99_{\pm3.3}$ & $91.54_{\pm1.9}$ & $94.26_{\pm2.9}$ \\
                               & Age Prediction  & $72.90_{\pm4.6}$ & $\underline{87.75}_{\pm2.0}$ & $64.60_{\pm 3.6}$ & $72.07_{\pm 3.0}$ & $92.07_{\pm2.7}$& $\underline{96.37}_{\pm0.9}$ \\
                               & Age Prediction w/ D.A & $74.06_{\pm3.4}$ & $86.90_{\pm1.6}$ & $\mathbf{65.79}_{\pm 2.0}$ & $\underline{73.02}_{\pm 4.3}$ & $\mathbf{94.01_{\pm{1.4}}}$ & $96.10_{\pm3.0}$ \\
                               & Age-Aware Contrastive Learning (ours) & $\textbf{76.33}_{\pm2.3}$ & $\textbf{88.11}_{\pm1.5}$ & $\underline{65.36}_{\pm3.7}$ & $\mathbf{73.33_{\pm4.3}}$ & $\underline{93.87}_{\pm1.3}$ & $\mathbf{96.84_{\pm2.3}}$ \\
     \hline
    \end{tabular}
    }
    \caption{Fine-tuning results using \rebuttal{100 or 500 (300 for AD vs HC) training subjects}. For each task, we report the AUC (\%) of the fine-tuned models initialized with different approaches with 5-fold cross-validation. 
    For age prediction, we employ the same transformations as in contrastive learning for the \rebuttal{Data Augmentation (D.A)} strategy. 
    Best results are in \textbf{bold} and second bests are \underline{underlined}.}
    \label{fine-tuning_table}
    \end{table}

\section{Conclusion}

Our key contribution is the introduction of a new contrastive 
loss, which leverages continuous (and discrete) meta-data from medical images in a self-supervised setting. We showed that our model, pre-trained with a large heterogeneous brain MRI dataset ($N=10^4$) of healthy subjects, outperforms the other SOTA methods on three binary classification tasks. In some cases, it even reaches the performance of a fully-supervised network without fine-tuning. This demonstrates that our model can learn a meaningful and relevant representation of healthy brains which can be used to discriminate patients in small data-sets. An ablation study showed that our method consistently improves upon SimCLR for three different sets of transformations. We also made a step towards a debiased algorithm by demonstrating that our model is less sensitive to the site effect than other SOTA fully supervised algorithms trained from scratch. We think this is still an important issue leading to strong biases in machine learning algorithms and it currently leads to costly harmonization protocols between hospitals during acquisitions. Finally, as a step towards reproducible research, we made our code public and we will release the BHB dataset to the scientific community soon.\\
Future work will consist in developing transformations more adapted to medical images in the contrastive learning framework and in integrating other available meta-data (\textit{e.g} participant's sex) and modalities (\textit{e.g} genetics). Finally, we envision to adapt the current framework for longitudinal studies (such as ADNI). \\

\textbf{Acknowledgments:} This work was granted access to the HPC resources of IDRIS under the allocation 2020-AD011011854 made by GENCI.

%
%
%
\bibliographystyle{splncs04}
\bibliography{bibliography_short}
\newpage
\appendix
\justify
\underline{\textbf{1. Demographic Information for BHB and UMAP visualization}}
\label{supp_bhb}
\begin{figure}[h!]
    \centering
     \begin{minipage}[c]{.40\linewidth}
        \centering
        \resizebox{\linewidth}{!}{
        \begin{tabular}{|c|c|c|c|c|c|c|c|}
            \hline
            Source & \# Subjects & N & Age & Sex (\%F) & \# Sites \\
            \hline
            \href{https://www.humanconnectome.org/study/hcp-young-adult}{HCP} & 1113 & 1113 & $29 \pm 4$ & 45 & 1 \\
            \href{http://brain-development.org/ixi-dataset}{IXI} & 559 & 559 & $48 \pm 16$ & 55 & 3\\
            \href{https://www.nitrc.org/projects/fcon_1000/}{CoRR} & 1371 & 2897 & $26 \pm 16$ & 50 & 19\\
            \href{https://openneuro.org/datasets/ds002330/versions/1.1.0}{NPC} & 65 & 65 & $26 \pm 4$ & 55 & 1 \\
            \href{https://openneuro.org/datasets/ds002345/versions/1.0.1}{NAR} & 303 & 323 & $22 \pm 5$ & 58 & 1 \\
            \href{https://openneuro.org/datasets/ds002247/versions/1.0.0}{RBP} & 40 & 40 & $23 \pm 5$ & 52 & 1 \\
            \href{https://www.oasis-brains.org}{OASIS 3} & 597 & 1262 & $67\pm 9$ & 62 & 3 \\
            \href{https://dataverse.harvard.edu/dataset.xhtml?persistentId=doi:10.7910/DVN/25833}{GSP} & 1570 & 1639 & $21 \pm 3$ & 58 & 1\\
            \href{https://ida.loni.usc.edu}{ICBM} & 622 & 977 & $30 \pm 12$ & 45 & 3\\
            \href{http://fcon_1000.projects.nitrc.org/indi/abide}{ABIDE I} & 567 & 567 & $17 \pm 8$ & 17 & 20\\
            \href{http://fcon_1000.projects.nitrc.org/indi/abide}{ABIDE II} & 559 & 580 & $15 \pm 9$ & 30 & 17\\
            \href{http://brainomics.cea.fr/localizer/localizer}{Localizer} & 82 & 82 & $25 \pm 7$ & 56 & 2\\
            \href{https://openneuro.org/datasets/ds000221/versions/00002}{MPI-Leipzig} & 316 & 316 & $37 \pm 19$ & 40 & 2\\
            \hline 
            \hline
            Total & 7764 & \textbf{10420} & $32 \pm 19$ & 50 & 74 \\
            \hline
        \end{tabular}
        }
        \caption{Demographic information}
        \label{tab:demographic_infos}
    \end{minipage}
    \begin{minipage}[c]{.59\linewidth}
        \centering
        \includegraphics[width=\textwidth]{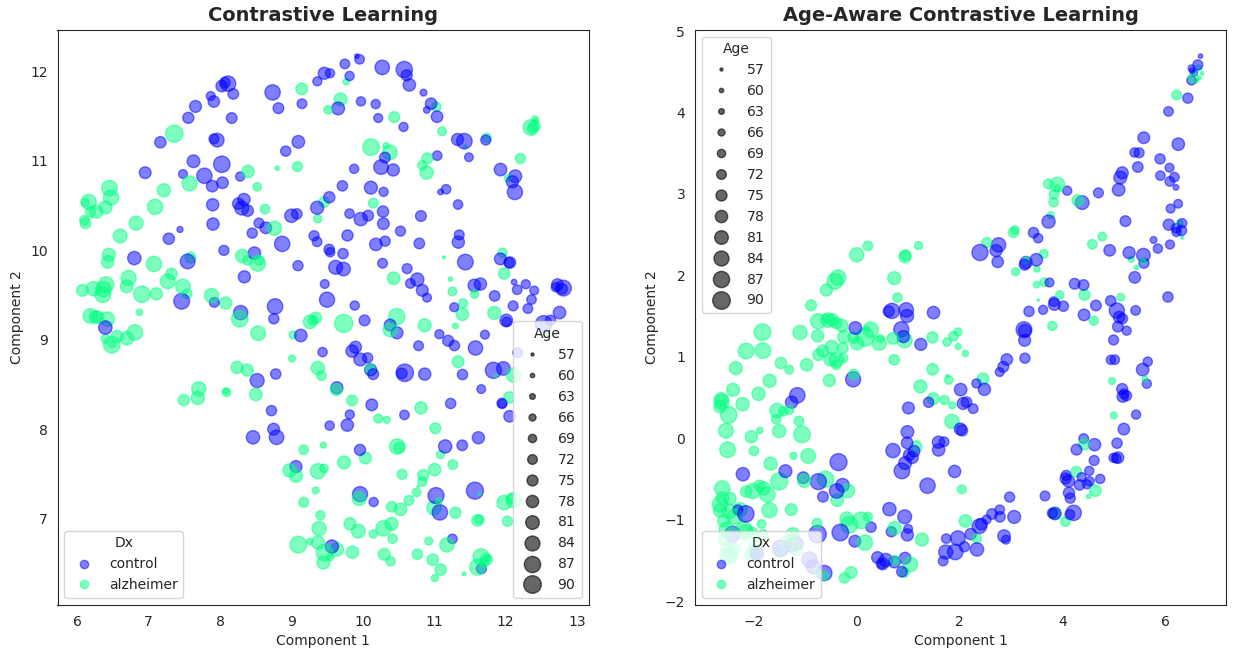}
        \caption{UMAP Representation}
        \label{fig:2D_umap}
    \end{minipage}
    
    \caption*{Fig. \ref{tab:demographic_infos}: Datasets aggregated for the BHB dataset. $N$ is the number of scans that passed the QC. Note that there might be several sessions (scans) per subject. Fig. \ref{fig:2D_umap}: 2D UMAP of ADNI features encoded (left) with SimCLR pre-training; (right) with our method. MRI from healthy participants with approximately the same age are mapped to the same region with our model. }
\end{figure}
\justify
\underline{\textbf{2. Hyperparameter selection: Batch Size and Patch Size for Cutout and Crop}}
\label{supp_hyperparams}



Fig. \ref{batch_size_tab_supp} (resp. \ref{patch_size_tab_supp}) we reported the performance of our model pre-trained with varying batch size (resp. black patch size for cutout and crop size) and $\sigma=5$. We performed a 5-fold CV under the linear evaluation protocol (encoder $f_\theta$ frozen) and we set $N_{target}=500$ for SCZ vs HC and BIP vs HC and $N_{target}=300$ for AD vs HC. We find that a large batch size is not necessarily required when dealing with brain MRI, in line with \cite{chaitanya2020contrastive}. We also fixed $p=25\%$ and $p'=75\%$ in our study.

\begin{figure}[h]
    \begin{minipage}[c]{.4\linewidth}
        \centering
        \begin{tabular}{|c|c|c|c|}
            \hline
            \multirow{2}{*}{\textbf{Batch Size}} & \multicolumn{3}{c|}{\textbf{Target Task}} \\
            \cline{2-4}
                                                 & SCZ vs HC & BIP vs HC & AD vs HC \\
            \hline
            64 & $82.94_{\pm 2.7}$ & $70.36_{\pm 2.6}$ & $93.03_{\pm1.8}$ \\ 
            100 & $84.15_{\pm 2.7}$ & $70.42_{\pm 1.1}$ & $93.53_{\pm1.6}$ \\
            \hline
        \end{tabular}
        \caption{AUC score (\%) as we vary the batch size during pre-training.}
        \label{batch_size_tab_supp}
    \end{minipage}
    \hfill%
    \begin{minipage}[c]{.4\linewidth}
        \centering
            \begin{tabular}{|c c|c|c|c|}
                \hline
                \multicolumn{2}{|c|}{\multirow{2}{*}{\textbf{Transformations}}} & \multicolumn{3}{c|}{\textbf{Target Task}} \\
                \cline{3-5}
                                                    & & SCZ vs HC & BIP vs HC & AD vs HC  \\
                \hline
                \multirow{2}{*}{Cutout} & $p=25\%$ & $82.94_{\pm2.7}$ & $70.36_{\pm2.6}$ & $93.03_{\pm1.8}$ \\ 
                                        & $p=50\%$ & $84.00_{\pm2.1}$ & $68.96_{\pm2.2}$ &  $89.21_{\pm2.7}$\\
                \hline
                \multirow{2}{*}{Crop} & $p'=75\%$ & $84.73_{\pm0.7}$ & $69.77_{\pm4.3}$ & $94.88_{\pm2.7}$\\ 
                                      & $p'=50\%$ & $81.77_{\pm3.1}$ & $68.69_{\pm 1.3}$ & $91.46_{\pm3.2}$\\
                \hline
            \end{tabular}
        \caption{AUC score (\%). The black patch size $p$ (for random cutout) and the crop size $p'$ are set during pre-training.} 
        \label{patch_size_tab_supp}
    \end{minipage}
\end{figure}

\justify
\underline{\textbf{3. Histogram of colors for MR images}}
\label{supp_color_histo}

\begin{figure}[H]
\setlength{\tabcolsep}{.2pt}
\begin{tabular}{m{.2\linewidth} >{\centering\arraybackslash}m{.4\linewidth} >{\centering\arraybackslash}m{.4\linewidth}}
    Original Image & Random Crop & Random Cutout \\
\end{tabular}
\includegraphics[width=\linewidth]{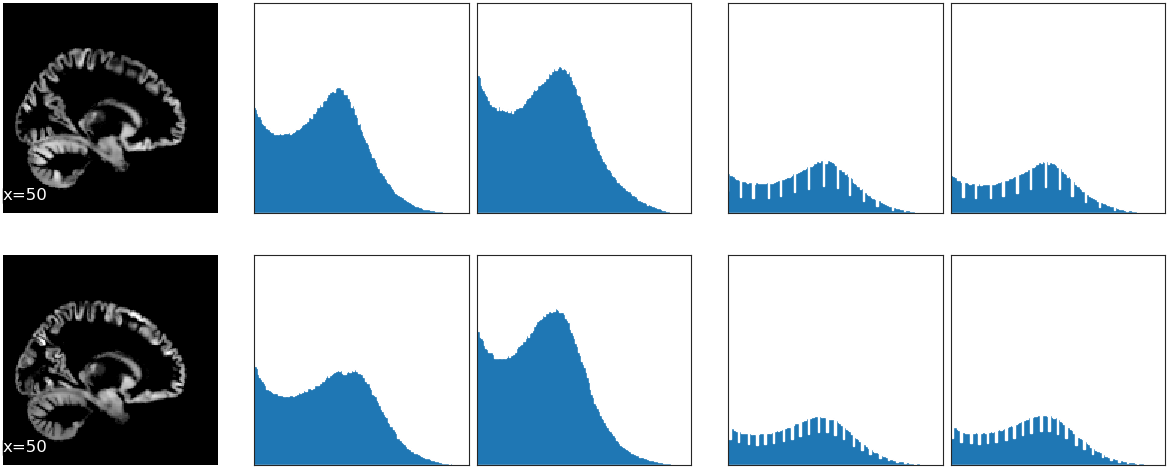}
\caption{Histogram of pixel intensities for 2 different images either i) randomly cropped or ii) partially masked with random cutout. We do not observe strong differences between the histograms for a given transformation. As such, color distortion may not be as critical as in \cite{chen2020simCLR}  to learn a robust representation since the network cannot take a shortcut based only on the color histogram.}
\label{color_histogram_crop_cutout}
\end{figure}

\justify
\underline{\textbf{4. Cross-Validation Strategies}}
\label{supp_CV}
\begin{figure}[h!]
    \centering
    \includegraphics[width=\linewidth]{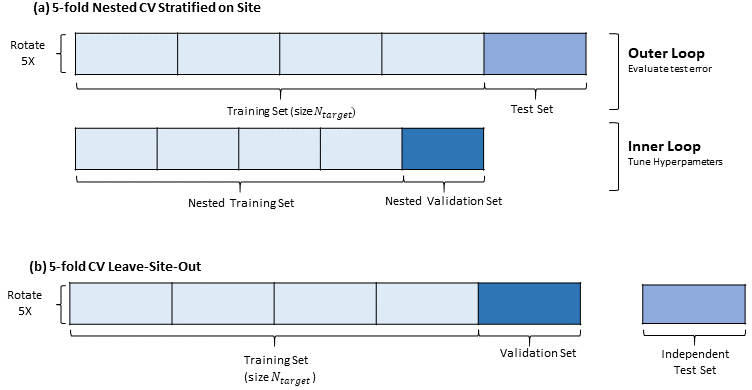}
    \caption{Cross-Validation Strategies used when evaluating our representation on the 3 downstream tasks. We both tested (a) and (b) with varying $N_{target}$.}
\end{figure}

\end{document}